\documentclass[10pt,journal,cspaper,compsoc]{IEEEtran}

\usepackage[nocompress]{cite}
\usepackage[pdftex]{graphicx}
\usepackage[cmex10]{amsmath}
\usepackage{url}
\usepackage[tight,normalsize,sf,SF]{subfigure}
\usepackage{multirow}

\ifx\oncomments\undefined\def\oncomments{1}\fi 
\usepackage{color}
\newcommand{\comments}[1] {\if\oncomments1 \textcolor{red}{ (#1) }\fi}

\begin{document}
\bstctlcite{IEEEexample:BSTcontrol}

%
\title{Emotional Storytelling using Virtual and Robotic Agents}

\author{Sandra~Costa,~\IEEEmembership{Member,~IEEE,}
        Alberto~Brunete,~\IEEEmembership{Member,~IEEE}\\
        Byung-Chull~Bae,~\IEEEmembership{Member,~IEEE}
        and~Nikolaos~Mavridis,~\IEEEmembership{Member,~IEEE,}

\IEEEcompsocitemizethanks{
\IEEEcompsocthanksitem S. Costa is with the Department
of Industrial Electronics Engineering, University of Minho, Portugal\protect\\
E-mail: scosta@dei.uminho.pt
\IEEEcompsocthanksitem A. Brunete is with the Center for Automation and Robotics (CAR UPM-CSIC), Universidad Politecnica de Madrid, Spain\protect\\
E-mail: alberto.brunete@upm.es 
\IEEEcompsocthanksitem B. Bae is with the School of Games, Hongik University, South Korea\protect\\
E-mail: byuc@hongik.ac.kr
\IEEEcompsocthanksitem N. Mavridis is with the Interactive Robots and Media Lab (IRML)\protect\\
E-mail: nmavridis@iit.demokritos.gr}
}


\IEEEcompsoctitleabstractindextext{%
\begin{abstract}
In order to create effective storytelling agents three fundamental questions must be answered: first, is a physically embodied agent preferable to a virtual agent or a voice-only narration? Second, does a human voice have an advantage over a synthesised voice? Third, how should the emotional trajectory of the different characters in a story be related to a storyteller's facial expressions during storytelling time, and how does this correlate with the apparent emotions on the faces of the listeners? The results of two specially designed studies indicate that the physically embodied robot produces more attention to the listener as compared to a virtual embodiment, that a human voice is preferable over the current state of the art of text-to-speech, and that there is a complex yet interesting relation between the emotion lines of the story, the facial expressions of the narrating agent, and the emotions of the listener, and that the empathising of the listener is evident through its facial expressions. This work constitutes an important step towards emotional storytelling robots that can observe their listeners and adapt their style in order to maximise their effectiveness.

\end{abstract}

}

\maketitle
\IEEEdisplaynotcompsoctitleabstractindextext
\IEEEpeerreviewmaketitle

\section{Introduction}
\label{Introduction}

Storytelling is a form of oral communication with pre-historic beginnings, which serves as a means of acculturation as well as transmission of human history \cite{roney1989back}. Traditionally, human storytelling has been one of the main means of conveying knowledge from generation to generation, but nowadays new technologies have also been used in this knowledge-sharing process\cite{chen2011survey}. 

A way to view the process of storytelling is the following: First, the storyteller understands the narrative message that is conceived by the story author. Second, the storyteller delivers it to the listener in an effective way. Unlike written narrative communication, however, where the author communicates with the reader through the implicit communication channel (real author -$>$ implied author -$>$ narrator -$>$ narratee -$>$ implied reader -$>$ real reader) \cite{lee2006attention}, the storyteller performs storytelling face-to-face in real-time. Thus, the storyteller can infer whether the listener is paying attention to the story from the listener's responses or back channels such as verbal responses (e.g., acknowledgement tokens such as yeah, uh huh, or mm hm) \cite{drummond1993back} and non-verbal responses (e.g., head nodding, eye blinking, or smiles). When the negative backchannels (e.g., head down or blank expression from boredom throughout the storytelling) are continuously recognised, an effective storyteller will change his or her narration technique to capture the listener's attention. While the storyteller's narration techniques will be various depending on the listener profiles (e.g., age, education, preferences, etc.), emotional expression (either verbal or non-verbal) is a common quality of the effective storyteller. 

The reader's emotional responses while reading a story in text can be either internal or external \cite{oatley1995taxonomy}. Examples of ``internal'' emotions (according to \cite{oatley1995taxonomy}) are identification or empathy with the story characters which occurs when the reader enters the story world. Examples of ``external'' emotions include curiosity and surprise which occurs when the reader meets with the narrative discourse structure through text \cite{oatley1995taxonomy}. In the same vein, the storyteller can elicit the listener's emotional responses either internally or externally. Specifically to arouse the listener's internal emotional responses, the storyteller can play an emotional surrogate of the characters or the narrator in the story. To enhance the listener's external emotional responses, the storyteller can pretend the listener's desirable emotional states (e.g., pretending curious or surprise more or less in an exaggerated way). 

The storyteller agent or system can detect the listener's non-verbal responses using various sensor devices. The detection of positive back channels is an indication of the listener's satisfaction or engagement in the story. In this case the storyteller system will continue to tell the story with the current storytelling rhetoric. If some negative back channels over the specified threshold are detected, however, this might be a sign of the listener's disliking or inattentiveness.

We have mentioned the word ``empathy'' before, a word that has a very important role in this paper. According to \cite{levenson1992empathy}, empathy is the ability to detect how another person is feeling, while \cite{decety2004functional} has defined empathy as: ``Empathy accounts for the naturally occurring subjective experience of similarity between the feelings expressed by self and others without losing sight of whose feelings belong to whom''. Empathy involves not only the affective experience of the other person's actual or inferred emotional state but also some minimal recognition and understanding of another's emotional state". Thus, empathy plays an important role in effective storytelling, as through observation of the listener's emotions, the storyteller can modulate his storytelling manner, in order to maximise effectiveness.

The storyteller has usually been a person, but recently some initial experiments with avatars as well as robots have taken place. If one wants to have humanoid robots in the role of learning companions and interaction partners to humans, engaging storytelling is a very important skill for them, according to \cite{cabibihan2012human}. 

Motivated by the above state of affairs, and towards our goal of effective emotional storytelling using Robotic and Virtual Agents, we performed two specially designed studies. In the first study we employed a virtual embodied conversational agent (Greta) as a storyteller that telling a story with emotional content expressed both by sound and facial expressions. In the second study we used a full-size highly-realistic humanoid robot (Aesop) with the same story material. In this paper we provide some initial empirical results of our studies, during which we observed the reactions of experimental subjects to artificial storytellers, through a combination of instruments: specially designed questionnaires, manual body language annotation, as well as automated facial expression and eye blink analysis.


\subsection{Research Questions and Expectations}
\label{RQ}

In this study, towards our ultimate purpose of effective emotional storytelling, we formulate three research questions (RQ1-RQ3) as follows:


\begin{itemize}
\item RQ1: Can a physical robot elicit more attention when telling a story instead of a virtual agent? (choice of embodiment for storytelling agent).
\item RQ2: While a story is told by the physical robot, will the participants show more non-verbal responses to a human voice recording as compared to a TTS voice recording? (choice of voice for storytelling agent).
\item RQ3: Will the participants empathise with the emotions conveyed by the robot as story narrator? (effectiveness of affective performance of storytelling agent).
\end{itemize}


\indent In order to answer RQ1 (effect of physical embodiment in listener attention), we measure attention through the eye blink rate of the listener, as it is well known that through parasympathetic mechanisms blink rate correlates to attention. In other words, spontaneous blinking rates are different depending on the types of behaviour (e.g., conversation \textgreater rest \textgreater reading \cite{bentivoglio1997analysis}) or visual information \cite{shultz2011inhibition}.) We expect that the audience blink rate will be less for the case of the physical robot storyteller (Aesop) , as compared to the the virtual agent storyteller (Greta). 

We furthermore expect attention to vary throughout the story, and the climax of attention to be at the climax of the story. 
Thus, apart from our main question RQ1 which was related to the difference of listener attention across embodiments (robot vs avatar), we also formulate a second subquestion - RQ1a: is there any difference between the various temporal parts of the story regarding blinking rate? RQ1 and RQ1a are addressed in experimental study 2 and study 1 respectively.

\indent Regarding RQ2 (effect of choice of synthetic vs. real voice), we codify the participant body language, expecting a real human voice to show signs of greater engagement as compared to the synthetic. This question is addressed in experimental study 2. 

\indent Finally, regarding RQ3, we compare empathy by measuring the similarity between the emotion line of the story and the emotion line of the observed facial expressions of the participants. In every story, we assume that there are different implied emotional lines over time for the characters, the narrator, and the listener. The listener's expected reaction is not the same as the narrators; rather, it is a function of all the above emotional lines. For example, the listener might feel anger about a sad character, because he or she might feel the situation is unfair. Our expectation is that the participants will empathise more with the story when the physical robot is the storyteller. 

 In the next Section (~\ref{Background}) relevant background literature using robots or virtual agents in storytelling scenarios is presented. Section~\ref{System} presents the architecture of our system, and Sections~\ref{Study1} and \ref{Study2} feature the procedures and results in the experiments of Studies 1 and 2. Section~\ref{Discussion} contains the discussion of the results, and conclusions are presented in Section~\ref{Conclusions}.

\section{Background}
\label{Background}

\subsection{Storytelling using Virtual Agents}
\label{virtual}

Several studies on emotionally expressive storytelling have been conducted using virtual agents. However, none have used automated analysis of non-verbal behaviour of the listeners with systems such as SHORE \footnote{http://www.iis.fraunhofer.de/en/bf/bsy/produkte/shore/} and faceAPI \footnote{http://www.faceapi.com/}, as we do in this paper. For example, Silva et al. \cite{silva2001papous}, \cite{silva2003tell} presented Papous, a virtual storyteller using a synthetic 3D person model with affective speech and affective facial/body expression. Papous could express six basic emotions (joy, sadness, anger, disgust, surprise, and fear) with different emotional intensity in the input text. The emotion tagging in the input script was simply made just for the narrator (i.e., storyteller) without considering possible emotions from different story characters or the listeners. 
\\ \indent In the CALLAS (Conveying Affectiveness in Leading-edge Living Adaptive Systems), an FP6 European research project, an integrated affective and interactive narrative system using a virtual character (Greta) was presented \cite{charles2007affective}. The proposed narrative system could generate emotionally expressive animation using a virtual character (Greta) and could adapt a given narrative based on the detected user emotions. In their approaches, the listeners's emotional states could be detected through emotional speech detection, which were applied to their interactive narrative system as either positive or negative feedback.
\\ \indent As a part of CALLAS project, Bleackley et al.\cite{bleackley2010emotional} investigated whether the use of an empathic virtual storyteller could affect the user's emotional states. In their study, each study participant was listened to a broadcast news about earthquake disaster twice - first, with only the voice along with relevant text, image, and some music; next, with an empathic virtual storyteller (Greta) along with the same conditions as the first. The SAM (Self Assessment Mannequin) test was used to measure the possible change of the user's emotional states in terms of PAD (Pleasantness, Arousal, and Dominance) level before and after listening to the news story. The results showed that the use of empathic virtual storyteller influenced the user's emotional states - the participants' average level of pleasantness and dominance were decreased respectively when the Greta was used as a proxy of empathic virtual storyteller. Our study on virtual storyteller was inspired by this mock-up study but we employed a fictional story with multiple characters and various narrative emotions (e.g. happiness, sadness, surprise, etc.) in it, and furthermore, we employ automated as well as  manual analysis of a both face and whole-body non-verbal behaviour.
former controls the story logic (such as story flow and coherency) and the latter keeps track of the reader's anticipated emotions. 
The notion of tracing the reader's anticipated emotions has some superficial similarity with our approach, but ours is focused rather on the listener's attention recognition and storyteller's empathising with the emotions of the story characters.
\\ \indent In order to measure the listener's attention to the story in an objective way, various factors (e.g., glancing, standing, nodding, or smiling) can be used to evaluate attentive engagement of the listener using visual information \cite{lee2006attention}. In our study we included eye blinking as a measurement of the listener's attention level based on the empirical studies claiming that inhibition of blinking is closely related to the intent of not losing important visual information \cite{bentivoglio1997analysis}, \cite{shultz2011inhibition}. For example, according to Bentivoglio et al.\cite{bentivoglio1997analysis}, eye blink rate shows a tendency of decreasing while reading (which requires more attention) and increasing during conversation (which requires less attention).

\subsection{Storytelling using Robots}
\label{robots}

Personal Electronic Teller of Stories (PETS) is a prototype storytelling robot to be used with children in rehabilitation  \cite{plaisant2000storytelling}. This robot was used remotely by children using a variety of body sensors adapted to their disability or rehabilitation goal. The children were meant to teach the robot how to act out different emotions such as sadness, or happiness, and afterwards to use storytelling software to include those emotions in the stories they wrote. The authors believed this technology was a strong encouragement for the children to recover quickly and may also help the children learn new skills. PETS' authors focused on guidelines for cooperation between adult and children, and design of game scenarios.
\\ \indent The ASIMO robot was used in a storytelling study where the goal was to verify how human gaze can be modelled and implemented on a humanoid robot to create a natural, human like behaviour for storytelling. 
The experiment performed in \cite{mutlu2006storytelling} provides an evaluation of the gaze algorithm, motivated by results in the literature on human-human communication suggesting that more frequent gaze toward the listener should have a positive effect. The authors manipulated the frequency of ASIMO's gaze between two participants and used pre and post questionnaires to determine the participants' evaluation of the robot's performance. The results indicated that the participants who were given more frequent gaze from ASIMO performed better in a recall task \cite{mutlu2006storytelling}. 
\\ \indent The GENTORO system used a robot and a hand held projector for supporting children's story creation and expression. Story creation included a script design, its visual and auditory rendering, and story expression as a performance of the script. The primary goal of the presented study was to clarify the effects of the system's features and to explore its possibilities. Using post-experimental questionnaires answered by the children, the authors affirmed that children had considerable interest in the robot, because it behaved like a living thing and always followed a path on a moving projected image \cite{sugimoto2011mobile}.
\\ \indent Pleo, a robotic dinosaur toy, has been used to mix physical and digital environments to create stories, which were later programmed with the goal of controlling robotic characters. Children created their stories, and programmed  how the robotic character should respond to props and to physical touch. 
The system gave children the opportunity and control to drive their own interactive characters, and the authors affirmed they contributed to the design of multimodal tools for children's creative storytelling creation \cite{ryokai2009children}.
\\ \indent In \cite{ribeiro2009robotics}, Lego Mindstorms robotics kits were used with children to demonstrate that robots could be a useful tool for interdisciplinary projects. The children constructed and programmed the Lego robots, addressing the dramatisation of popular tales as the final goal. The study results showed the applicability of robots as an educational tool, using storytelling as a background, developing thinking, interaction and autonomy in the learning process.
\\ \indent The results of two-year's research in a classroom of children with intellectual disabilities and/or autism are described in \cite{munekata2009new}. The PaPeRo robot was used to enhance storytelling activities. The authors found that the length of stories produced by the children continually increased and the participants began to tell more grammatically complex stories. 
\\ \indent From \cite{chen2011survey}, a survey on storytelling with robots and other simple projects are presented. Summarising, the main users or the target group of the presented studies are normally developed \cite{sugimoto2011mobile} or disabled children \cite{plaisant2000storytelling}, either with the goal of teaching/learning or rehabilitation. Adults were involved in some projects \cite{mutlu2006storytelling}, but usually as teachers \cite{ribeiro2009robotics}. The robots used in the previous projects were mostly small mobile robots (e.g Lego Mindstorms or Pleo), and the outcomes of the studies were mainly design, pedagogics, and prototypes in authoring, learning or mixed environments.
\\ \indent Comparing our work with the studies presented above, in the study presented in this paper, the target group is composed only of adults. However, our methodology could also be used with children. So, we have a different and more general listener, and most importantly we are directing our goals towards the automated observation of the emotional and non-verbal communication provided by the participants while reacting to the story told by the robot or the virtual agent. One of the main novelties of our study thus consists in the analysis of the whole-body non-verbal communication, and the automated analysis of the facial expressions made by the participants during the storytelling experiments, as well as in the direct comparison between virtual avatar and real android robot embodiments.


\section{System Design}
\label{System}

In this section the architecture of the system is described. The system employs verbal and non-verbal rhetoric (e.g., emotional speech and facial expression) as a way of empathising storytelling. listener attention and emotion recognition is performed using special automated software analysers (FaceAPI and SHORE), in conjunction with standardised human observation and formalised description (for the case of non-facial body language).

\subsection{Overall Architecture}

\begin{figure}[tb]
\centering
\includegraphics[width=0.45\textwidth]{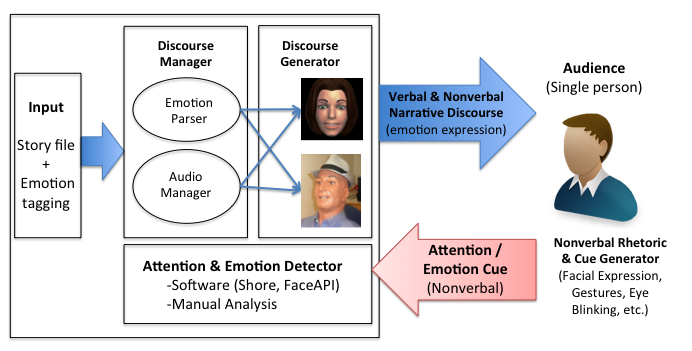}
\caption{System architecture.}
\label{Fig1}
\end{figure}

Two embodiments of the storytelling agent are utilised in this paper: a virtual agent (Greta \cite{poggi2005greta}), and a physical humanoid robot (Aesop, a version of the Ibn Sina robot described in \cite{mavridis2009ibnsina}, \cite{riek2010ibn}, \cite{mavridis2012opinions}).
\indent  As illustrated in Fig. \ref{Fig1}, our proposed system consists of three main components: Discourse Generator, Discourse Manager and Attention and Emotion Detector. The input text story file was manually tagged with possible emotions and was formatted using FML-APML (Function Markup Language - Affective Presentation Markup Language) \cite{mancini2008fml}. FML-APML is an XML-based markup language for representing the agent's communicative intentions and the text to be uttered by the agent. This version encompasses the tags regarding emotional states which were used to display different emotions both on the virtual and on the robotic agents.

\subsection{Discourse Manager}

Discourse Manager consists of two modules: Emotion Parser and Audio Manager. Emotion Parser extracts emotion information from the input text file and formats it to be used by Greta or Aesop. The Audio Manager selects if the audio will be produced by a TTS systems or is provided as a pre-recorded voice.

\subsection{Discourse Generator}

This module creates the storytelling experience. There are four options depending on the selected agent (Greta or Aesop) and the audio generation (TTS or pre-recorded voice). For Greta, if TTS is chosen, both video (with facial expressions corresponding to emotions) and audio and generated by a computer. If pre-recorded voice is chosen, the video (with facial expressions) is merged with the provided audio file.

The Aesop physical humanoid robot, on the other hand, uses its facial expression capabilities to show the emotions provided by the emotion parser, while the voice comes from TTS or pre-recorded audio.

\subsection{Attention and Emotion Detector}

This module takes as input a video from an listener-observing camera in real-time and analyses the listener response in terms of emotions and attention. In this paper SHORE (Sophisticated High-speed Object Recognition Engine) \cite{shore} is used to recognise the listener emotional facial expressions, and FaceAPI \cite{faceAPI}, a real-time face tracking toolkit, to extract attention-related features (eye-blink) from the listener's face. 

\indent SHORE \cite{shore} has a face detection rate of 91.5\%, and the processing speed of full analysis including facial expressions is 45.5 fps. SHORE recognises the following facial expressions: Happy, Surprised, Angry, and Sad. The software is capable of tracking and analysing more than one face at a time in real-time with a very high robustness especially with respect to adverse lighting conditions.

\indent FaceAPI \cite{faceAPI} provides an image-processing modules for tracking and analysing faces and facial features. FaceAPI provides a real-time, automatic monocular 3D face tracking, and it tracks head-position in 3D providing X,Y,Z position and orientation coordinates per frame of video. FaceAPI also enables blink detection, and can also track 3 points on each eyebrow and 8 points around the lips.


\section{Experimental Study 1}
\label{Study1}
This study mainly addressed experimental questions RQ2, dealing with the effect of voice choice on listener engagement, and RQ3, focusing on the relation between the emotional line of the story and the emotional reactions of the listener.

\subsection{Procedures}
\label{ProceduresStudy1}
We adopted a short story titled ``The cracked pot'' (consisting of 12 sentences and about 250 words) as a story material. This story is based on a Chinese parable, and has a lot of attractive features for our study. First, it has the right length - neither too long, nor too short: a little more than two minutes (2:09 in our narration). Second, it has a main character (the cracked pot), which we expect that the listener will feel affection for. Third, it has an active emotional trajectory of intermediate complexity.

In the ``The cracked pot'' a heterodiegetic narrator (i.e., the narrator who is not present in the story world as a character) narrates a story consisting on three characters (a woman, a perfect pot, and a cracked pot) with omniscient point of view. 
\\ \indent In order to annotate the emotion line of the story, five adults individually tagged the possible character emotions sentence by sentence. The emotion category was limited to six basic emotions (Happiness, Sadness, Anger, Fear, Disgust, and Surprise) with intensity range from 0 (Not at all) to 10 (Extreme). The tagged data were collected and averaged with the confidence ratio based on the number of the responses (Resulting tagged emotions in Table \ref{tab:1}). 


\begin{table*}
\caption{Story material (The Cracked Pot) and tagged emotion intensity (H for Happiness, s for Sadness, Su for Surprise)}
\label{tab:1} 
\begin{tabular}{|p{14cm}||l|}
\hline
Sentence & Narrator  \\ 
\hline
\hline

An elderly Chinese woman had two large pots, each hung on the ends of a pole, which she carried across her neck(0s-10s).&  H 1(.2)\\[13pt]

One of the pots had a crack in it while the other pot was perfect and always delivered a full portion of water(10s-22s). & S 4(.4) H 2(.2)\\[13pt]

At the end of the long walk from the stream to the house, the cracked pot arrived only \textbf{half full} (22s-30s).& S 4.5(.8)\\[5pt]

For a full two years this went on daily, with the woman bringing home only one and a half pots of water (30s-39s). & S 5.5(.2)\\[13pt]

Of course, the perfect pot was proud of its accomplishments (39s-45s). & H 4.5(.4)\\[5pt]

But the poor cracked pot was ashamed of its own imperfection, and miserable that it could only do half of what it had been made to do (45s-53s). & S 8.5(.4)\\[13pt]

After 2 years of what it perceived to be bitter failure, it spoke to the woman one day by the stream (53s-62s).& S 6(.6) Su 1(.2)\\[5pt]

"I am ashamed of myself, because this crack in my side causes water to leak out all the way back to your house (62s-75s)." & S 7(.4)\\[13pt]

The old woman smiled, "Did you notice that there are \textbf{flowers} on your side of the path, but not on the other pot's side?" (75s-91s) & Su 4.5(.8)H 5.5(.4)\\[13pt]

"That's because I have always known about your flaw, so I planted flower seeds on your side of the path, and every day while we walk back, you water them."(91s-106s) & H 5.4(1.0) Su 5.5(.4)\\[13pt]

"For two years I have been able to pick these beautiful flowers to decorate the table (106s-113s).& H 7(.6)\\[5pt]

Without you being just the way you are, there would not be this beauty to grace the house (113s-125s)." & H 7.4(1.0)\\[1pt]
\hline
\end{tabular}
\end{table*}

Fig. \ref{Fig2} shows the emotion line dynamics obtained from the emotion tagging by our human annotators in which the intensity of each emotion was obtained considering the confidence ratio. For example, as seen in Fig. \ref{Fig2}, the emotion of surprise was present as a transition emotion from a negative emotional state (i.e., sadness) to a positive emotional state (i.e., happiness).

\subsubsection{Participants}
\label{ParticipantsStudy1}
A total of 20 participants (10 women, 10 men), who were students, staff, and researchers from New York University Abu Dhabi, were volunteer participants in experimental study 1. Their ages ranged from 18 to 60 years old. Each participant was arbitrarily assigned, while balancing the gender ratio, to one of the two groups. Ninety percent of the participants used English as a foreign language, while the others were native speakers. The participants in one group (Group A) listened (individually) to only the pre-recorded audio story which is narrated by a human storyteller (no embodiment whatsoever); the participants in the other group (Group B) listened (individually) to the same audio story with video in which Greta expressed her emotions using facial expressions.

\subsubsection{Experimental Setup}
\label{SetupStudy1}
The cracked pot story was recorded using an amateur female voice actor.
\\ \indent When participants entered the room where the study was conducted, they were guided to sit on a chair. A video camera was set up in front of the participants. The participants were given a SAM (Self Assessment Manikin) test sheet to describe their current emotional states and answered a pre-experiment questionnaire consisting of the questions about their demographic information. 
\\ \indent The participants in Group A listened to the audio story through the speaker in the room, without any other relevant material to the experiment; the participants in Group B watched a 50-inch TV screen on the wall in which Greta showed her emotional facial expressions according to the same audio story. Facial expressions of participants were recorded under their agreement. After the storytelling is over, the participants were asked to provide ratings on a 7-point scale, ranging from ''not at all'' (1) to ''very much'' (7) about their story appreciation. They were also asked to provide their opinions about the experiment as open questions. Finally they described their current emotional states using the same SAM test. The questions set in the experiment questionnaire included the following:\\
Q1. How interesting was the story?\\
Q2. How sorry did you feel for the bad pot in the beginning of the story?\\
Q3. How much did you enjoy listening to the story?\\
Q4. How much did you want to know how the story would end?\\
Q5. How happy did you feel for the bad pot at the end of the story?\\
Q6. How much did you like the story?\\
Q7. What emotions did you feel while listening to the story? (Please describe all the emotions you felt.)\\
Q8. What emotions did you notice while listening to the story? (Please describe all the emotions you noticed.)

\subsubsection{Evaluation Tools}

Besides standard statistical software to analyse the data produced from the experiments, SHORE \cite{shore} and faceAPI \cite{faceAPI} were used. SHORE was used for listener facial expression recognition. 
\\ \indent In particular for this study, it is important the recognition of facial expressions (happy, surprised, angry, and sad) and the detection of in-plane rotated faces (up to +/- 60 degree), in order to be able to examine the facial expressions displayed by the participants and their relation with the ones conveyed by the story, towards research question RQ3. Classification accuracy of basic emotions through facial expressions typically ranges between 85\% and 95\% \cite{pantic2000automatic}.
\\ \indent The second software used in parallel to SHORE for automated analysis, namely FaceAPI, provides functionality for tracking and analysing faces and facial features. It was used for eye-blinking detection.

\subsection{Results}
\label{ResultsStudy1}
None of the participants had to be excluded due to their performance in the SAM test. The mean ratings of the survey questionnaire of the participants in Group A (audio only) and the participants in Group B (the same audio but with video using Greta's emotional facial expressions) are shown in Table \ref{tab:2} in terms of the three story appreciation factors - liking, engagement, and empathising. 

\begin{table*}\centering
\caption{Comparison of the mean ratings between two groups in 7-point scale rating.}
\label{tab:2}
{\begin{tabular}{@{}|c|c|c|c|c|c|c|@{}} 
\hline
& \multicolumn{3}{c}{Audio Only (Group A)}  \vline & \multicolumn{3}{c}{Audio and Video (Group B)} \vline \\
\cline{2-7} 
& Male & Female & Overall & Male & Female & Overall \\
& M(SD)& M(SD)&M(SD)&M(SD)&M(SD)&M(SD)\\ 
\hline
Q1&3.8 (1.79)&5.0 (2.0)&4.4 (1.89)&4.6 (1.67)&4.6 (1.67)&4.6 (1.59)\\
Q2&2.2 (1.64)&4.0 (2.0)&3.1 (1.82)&4.4 (1.95)&4.2 (1.95)&4.3 (1.72)\\
Q3&3.2 (1.30)&4.8 (2.17)&4.0 (1.74)&3.8 (1.30)&4.4 (1.30)&4.1 (1.49)\\
Q4&4.0 (1.87)&5.6 (1.14)&4.8 (1.51)&5.0 (2.35)&5.2 (2.35)&5.1 (1.59)\\
Q5&4.4 (1.95)&6.4 (1.34)&5.4 (1.65)&4.8 (1.30)&5.4 (1.30)&5.1 (1.22)\\
Q6&4.2 (2.17)&5.6 (1.52)&4.9 (1.84)&4.6 (1.67)&4.8 (1.67)&4.7 (1.98)\\ 
Liking (Q3+Q6) &3.7 (1.77)&5.2 (1.81)&4.5 (1.90)&4.2 (1.48)&4.6 (1.67)&4.4 (1.67)\\
Engagement (Q1+Q4) &3.9 (1.73)&5.3 (1.57)&4.6 (1.76)&4.8 (1.93)&4.9 (1.20)&4.9 (1.57)\\
Empathizing (Q2+Q5) &3.3 (2.06)&5.2 (2.04)&4.3 (2.22)&4.6 (1.58)&4.8 (1.40)&4.7 (1.45)\\
\hline
\end{tabular}
}
\end{table*}

\begin{figure}[tb]
\centering
\includegraphics[width=0.5\textwidth]{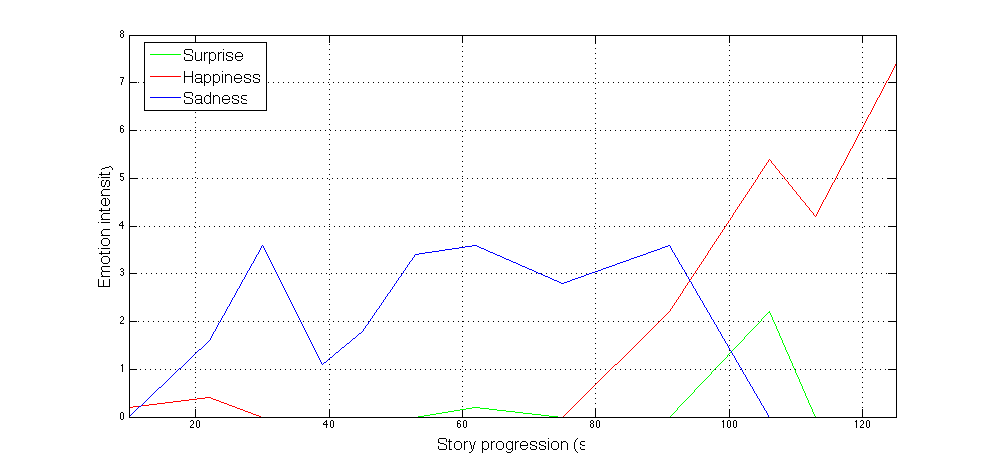}
\caption{The narrator's emotion dynamics in the cracked pot story.}
\label{Fig2}
\end{figure}

\begin{figure}[tb]
\centering
\includegraphics[width=0.5\textwidth]{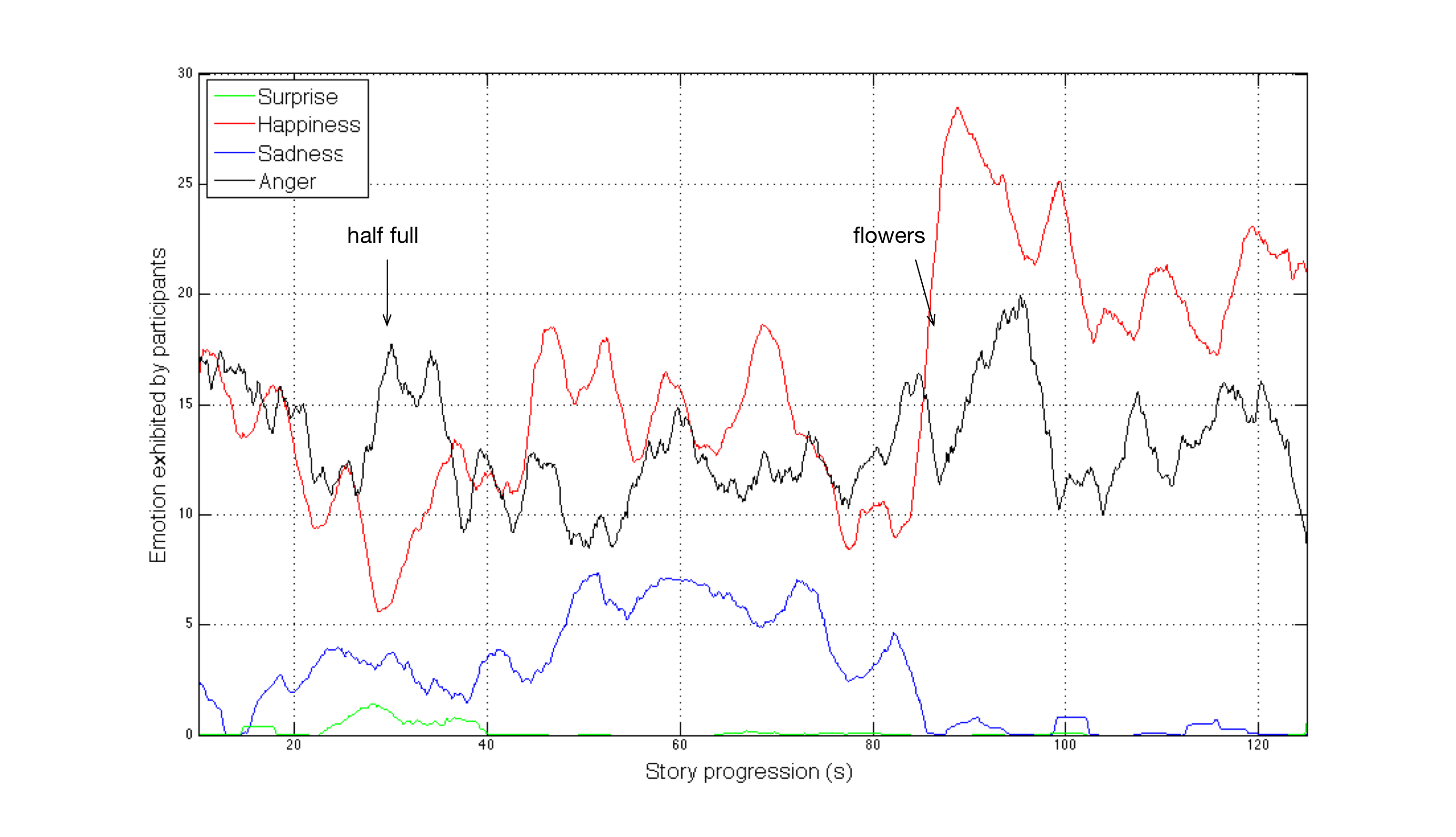}
\caption{Percentage of the emotions shown in participants during storytelling.}
\label{Fig3}
\end{figure}

The mean ratings of the overall survey questions and the three factors between two groups are not statistically significant (p \textgreater .05). However, when it comes to the gender difference, male participants prefer storytelling with visual stimulus. Particularly among the three factors, the mean ratings of the empathising questions (Q2 and Q5) in Group A are indeed significantly different: the mean ratings of male participants is 3.3 and those of female participants is 5.2 (p = .05). The female participants are relatively less affected by the visual stimulus.
\\ \indent Fig. \ref{Fig3} shows the emotional analysis of the listener, which was made through automated classification of facial expressions using the SHORE software. It shows the different emotions measured from the participants' faces while watching the video. A very interesting comparison here is with Fig. \ref{Fig2}, which contains the facial expressions that the storytelling agent (robot or avatar) was programmed to perform during storytelling, on the basis of the tagged emotions of table 1. Let us have a deeper look.

First, notice the overall synchronisation of the main transitions, between Fig. \ref{Fig2} and Fig. \ref{Fig3}. In Fig.3 two keywords have have been overlapped to facilitate the analysis. The storytelling timeline of Fig. \ref{Fig2} contains a number of important events E1-E3, on the basis of the observed emotional transitions: 

\begin{itemize}
\item E1: There is a sharp narrator sadness peak (blue line) that occurs around t=30s (almost in sync with the words "half full" in the story text), with a roughly triangular supporting ramp lasting between t=23s and t=37s.
\item E2: There is a second more stable narrator sadness plateau lasting roughly between t=45 and dominating till t=90 or so.
\item E3: Happiness takes over the narrator's facial expressions from t=90 approx. (almost in sync with the word "flowers" in the story text") until the end of the story.
\end{itemize}

Correspondingly, moving from the narrator emotional facial expressions in Fig. \ref{Fig2} to the resulting listener emotions as witnessed by automatically analysed facial expressions in Fig. \ref{Fig3}, one can notice the following events:

\begin{itemize}
\item E1': In good synchrony to E1, the sharp narrator sadness peak (blue line) produces a marked decrease in happiness and increase in anger in the listener (t=23s to t=37s)
\item E2': During the story period where the hero of the story (cracked pot) is sad and no positive signs appear on the horizon (and strong words such as "miserable", "bitter failure" are heard, the listener experiences increasing and then sustained sadness too, which is also the emotion the narrator expresses in E2 (t=45s to t=90s)
\item E3': Roughly when the word "flowers" is heard ("The old woman smiled, did you notice that there are flowers on your side of the path?"), there is a great increase in apparent happiness in the listener which after the first peak is sustained all the way to the end of the story, in response to E3 (t=90s until t=125s)
\end{itemize}

Thus, what is apparent is that:
\begin{itemize}
\item There exists synchrony between story content, narrator facial expressions, and resulting listener facial expressions.
\item The relation between these three timelines is not a simple equality or one-to-one relation, but contains both its own "harmonies" as well as "dynamics". By "harmonies" we are referring to the relations of the emotions across the implicated timelines. For example, during event E3, the happiness of the narrator is connected to the happiness of the listener in E3', and this is a simple equality relation (Narrator Happy - listener Happy). However, this is not the case in E1: there, the sadness of the narrator is reflected to anger in the listener; but this relation does not always hold: for example, the narrator sadness in E2 causes an increase in listener sadness in E2', and not anger as it did in E1. We will further discuss these important observations below.
\item The baselines (average value and scale) in the four emotional lines of Fig. \ref{Fig3} are different for each emotional component, and thus relative changes might well be a stronger indicator rather than absolute values. A more detailed discussion and elaboration will follow in the next sections.
\end{itemize}

\begin{figure}[tb]
\centering
\includegraphics[width=0.5\textwidth]{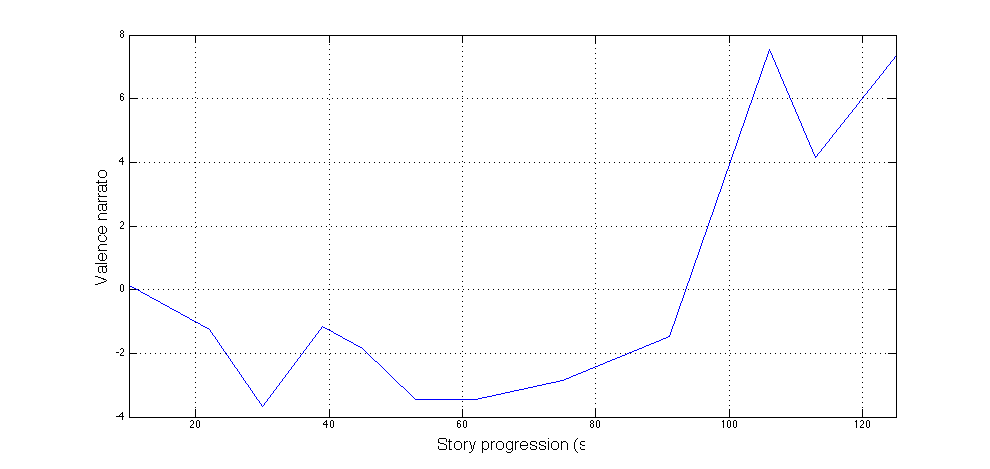}
\caption{Narrator valence}
\label{Fig3a}
\end{figure}

\begin{figure}[tb]
\centering
\includegraphics[width=0.5\textwidth]{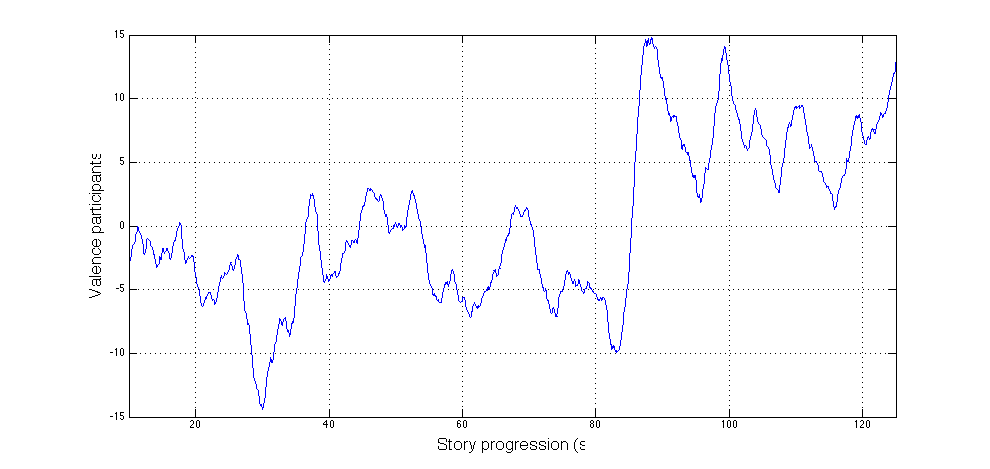}
\caption{Participants valence}
\label{Fig3b}
\end{figure}

It is well known from the literature that a primary dimension of emotional spaces is that of valence. Therefore, in order to better understand the results, we have decided to also analyse the valence of the narrator and the participants emotions. We have considered the formula in Eq. \ref{eq1}, where h stands for happiness, su for surprise, a for anger and s for sadness. m is the mean of each vector. It shows the relation between positive feelings (happiness and surprise) and negative feelings (sadness and anger).

\begin{equation}
\label{eq1}
•v[t]=(h[t]-m_h+su[t]-m_s-(a[t]-m_a+s[t]-m_s)
\end{equation}

Figures \ref{Fig3a} and \ref{Fig3b} show the results of the narrator and the participants respectively. It is possible to see that they have a similar pattern, meaning that in terms of valence (positive and negative feelings), the participant empathises with the narrator.

\begin{figure}[tb]
\centering
\includegraphics[width=0.4\textwidth]{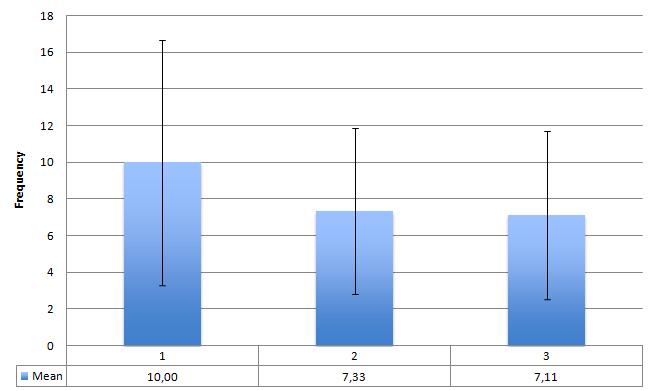}
\caption{Average number of eye blinking of the participants in subject group B during storytelling}
\label{Fig4}
\end{figure}

Now let us move on from facial expressions to eye blink rate. Fig. \ref{Fig4} shows the average blinking frequency of the listener during storytelling. 
The story has been divided in three parts to compare the number of eye blinks in each of them. It can be seen that as the story progresses the number
 of eye blinks decreases, signalling an increase in participants' attention as they got involved in the story. A t-test performed over this data shows 
 that the differences between eye blink rates between the story sections are indeed statistically significant (p \textless .05), providing evidence towards question RQ1a: i.e. our empirical data supports the hypothesis that as we are reaching towards the
  climax which occurs almost at the end of the story, apparent listener attention is increasing.

\section{Experimental Study 2}
\label{Study2}

As it was mentioned previously we wanted to verify the differences between a virtual and a robotic agent in a storytelling scenario, thus addressing research question RQ1. A new study with different participants was thus designed. In addition to this, special attention was now given also to the participants' non-facial body language, which was video-taped and annotated using a special formal scheme (used by the Observer XT 11 software by Noldus \cite{noldus1991observer}) that will be described.

\subsection{Procedures}
\label{Procedures}

Some elements used in Study 1 were used in Study 2, which are identified in the corresponding section, and the similarities and differences are pointed out. The same story as used in Study 1 was employed. 

\subsection{Participants}
\label{Participants}
A total of 40 (8F, 32M) students and staff of NCSR-D Research Center volunteered, with ages from 20 to 55 years (21 - 20/29 years old (YO); 14 - 30/39 YO; 4 - 40/49 YO; 1 - 50+). 
Each participant was arbitrarily assigned to one of the two groups. All used English as a foreign language.

\subsection{Experimental Setup}
\label{ExperimentalSetup}

The participants in Group A listened to the story with a human voice recording and the participants in the Group B listened to the same audio story with a voice produced by text-to speech software. Besides having Greta telling the story to the participants, Aesop, a humanoid robot told ``The cracked pot'' story. This robot was already used in other scenarios \cite{christoforou2010android}, \cite{mavridis2009ibnsina} but in this study it was moving only its head, and not its hands, displaying different facial expressions according to the emotion conveyed in the story.


In the experiments, the participant was welcomed into a room and he/she was requested to sit on a chair. The room setup includes a table, the robot at one side and the participant at the other, and two cameras on both sides of the robot. The participant was requested to fill in a consent form, and an online questionnaire, the same as used in Study 1. The SAM test was applied before and after the experiment, with the same goal as in Study 1.



All participants listened to the story in which Greta showed her emotional facial expressions accordingly. Half of the same participants also observed the Aesop robot narrating the story using a human voice and facial expressions, and the other half using a TTS voice and facial expressions. The order in which the participants listened to the story (either by Greta or by Aesop) was random.
\\ \indent When the storytelling was over, the participants were asked to provide ratings on a 7-point Likert scale ranging from ''not at all'' (1) to ''very much'' (7) about their story appreciation. Open questions were also asked to get opinions about the experiment.

\subsubsection{Evaluation Tools}
Besides the same evaluation tools used in Study 1, the videos produced during the experiments were examined using the Observer XT 11 program by Noldus \cite{noldus1991observer}. This software is a video annotation tool, and it was used to code events related with non-verbal body communication. Three different categories of behaviours were defined to be identified in the videos: hand and arm gestures, legs gestures, and head position. Inside these categories, the codes used were:

\begin{itemize}
\item Hand and Arm Gestures: steeping hands, evaluation, index finger, chin stroking, crossed arms; 
\item Legs Gestures: crossed legs, 4 leg lock, leg clamp;
\item Head Positions: neutral head position, interested position, disapproval position, hands behind head.
\end{itemize}

These events were chosen from \cite{pease2012body} having in mind participants' behaviours which would help us to answer RQ2 and RQ3. Summarily, some of these behaviours might help us to evaluate the non-verbal communication performed by the participants, indicating for example that the participants are interested, listening, evaluating, or disagreeing with the situation they are involved in. 

Observation of gesture and congruence of the verbal and non-verbal channels are the keys to accurate interpretation of body language. However, all gestures should be considered in the context in which they occur \cite{pease2012body}. Head orientation was suggested by Ekman and Friesen to be an indicator of gross affective state (positive/negative) as well as intensity of emotion \cite{Ekman:1967}. A user study using the PAD model to assess the perception of affect from head motion during affective speech reported that head motion corresponding to distinct affective states is described by different motion activation, range, and velocity. \cite{Busso:2007}. 

Several studies regarding hand and arm movements show significant results for distinguishing between affective states \cite{Wallbott:1998}\cite{Fast:1998}. These are recognized above chance level in full-light videos of hand and arm movements \cite{Carmichael:1937}\cite{Reilly:1992}, animated anthropomorphic and non-anthropomorphic hand models displaying abstract movements \cite{Samadani:2011}. 

To ensure inter-rater reliability 10\% of the videos were re-coded by a second independent coder (Cohen's kappa $k= .64$). This is acceptable, as having a Cohen's kappa value higher than 0.60 suggests a good agreement between the raters \cite{Bakeman:1997}. When the coders were analysing the behaviour of the listener, they observe the videos from both cameras simultaneously and were able to hear the whole interaction.


\subsection{Results}
\label{ResultsStudy2}

\subsubsection{Questionnaires} 
\label{Questionnaires2}
\begin{figure}[tb]
\centering
\includegraphics[width=0.5\textwidth]{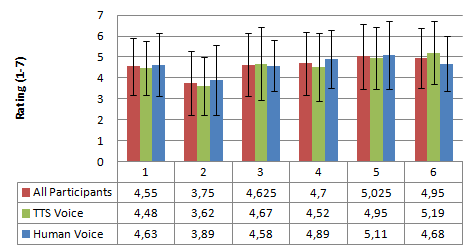}
\caption{Average of the questionnaire's ratings.}
\label{Fig7}
\end{figure}

None of the participants had to be excluded due to their performance in the SAM test. Fig. \ref{Fig7} presents the results from the questionnaires given to the participants. It shows the mean ratings of the participants' answers who heard the story told by Aesop using a TTS voice, using a human voice, as well as both. Between these two groups there are no significant differences regarding their ratings. However, in general, participants who listened to the story with the human voice rated the questions higher than the other participants. Questions are described in Section \ref{SetupStudy1}.

Table \ref{tab:3} presents the comparison of the three story appreciation factors - liking, engagement, empathising the ratings of participants who heard the story told by a human voice or a TTS voice.

\begin{table}[h]
\centering
\caption{Comparison of the mean ratings between two groups in 7-point scale rating - M(SD).}
\label{tab:3}
{\begin{tabular}{|l||l|l|l|}
\hline
& All participants & TTS Voice & Human Voice\\
\hline
Liking & 4.79 (1.46) & 4.93 (1.62)& 4.63 (1.25)\\
(Q3+Q6) &&&\\
Engagement & 4.63 (1.44) & 4.5 (1.46) & 4.76 (1.43)\\
(Q1+Q4) &&&\\
Empathizing & 4.39 (1.53) & 4.29 (1.45) & 4.5 (1.65)\\
(Q2+Q5) &&& \\ 
\hline
\end{tabular}
}
\end{table}

A one-way ANOVA revealed \textit{significant differences} when comparing the scores of Q2 and Q5. These questions were related with the empathy generated between the participant and the storyteller (F(39, 77) = 13.90; p \textless .001). In the first question (Q2) the participants indicated if they felt sorry about the main character (beginning of the story), and the second question (Q5) if they felt happy (end of the story). The average of the ratings in these two questions indicate that at the end of the story the participants were more engaged with the story character than in the beginning. Since this questionnaire was done at the end of the experiment, when all the participants listened to the story both by Greta and by Aesop, through this questions alone, it is not possible to infer which one was a more effective storyteller. 
\\ \indent A one-way ANOVA test showed no significant differences when comparing Q1 with Q4 (F(39, 77) = 0.643; p =  .22) and Q3 with Q6 (F(39, 77) = 0.321; p =  .99). The pairs Q1/Q4, and Q3/Q6 were connected to the same appreciation factor, and thus it makes sense that no significant differences were found. Thus, our main finding here is that at the end of the story the participants reported feeling happy in a statistically significant stronger way as compared to their subjective reports of feeling sorry near the beginning of the story. This could be explained in multiple ways: either the magnitude of the sorry feeling was anyway smaller than the happiness at the end, or there is a memory effect which diminishes the remembered intensity of feeling sorry at the time of questionnaire answering, which takes place after the sorry feeling has been replaced by happiness. Also, and quite possibly, the empathy level of the listener towards the hero (the cracked pot) has increased through the progression of the story and the listener's familiarisation with the hero, and this contributes positively to the intensification of the subjective reports of happiness, which is also the final emotional state of the story - in a sense, as the saying goes, all is well that ends well.

\subsubsection{Emotional Analysis}
\label{emotionalAnalysis2}

\begin{figure}[tb]
\centerline{
	\subfigure[Emotions shown in participants during storytelling with Aesop with a human voice.]{
		\label{Fig13_1}
		\includegraphics[width=0.4\textwidth]{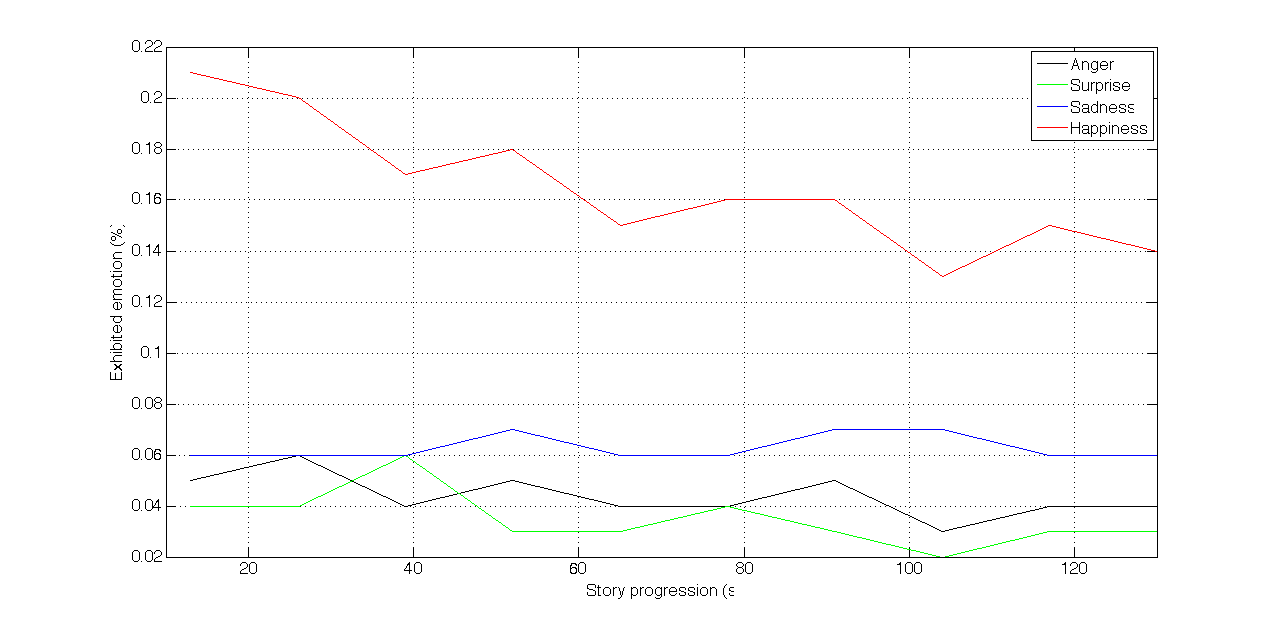}
	}
}\centerline{
	\subfigure[Emotions shown in participants during storytelling with Aesop with a TTS voice.]{
		\label{Fig13_2}
		\includegraphics[width=0.4\textwidth]{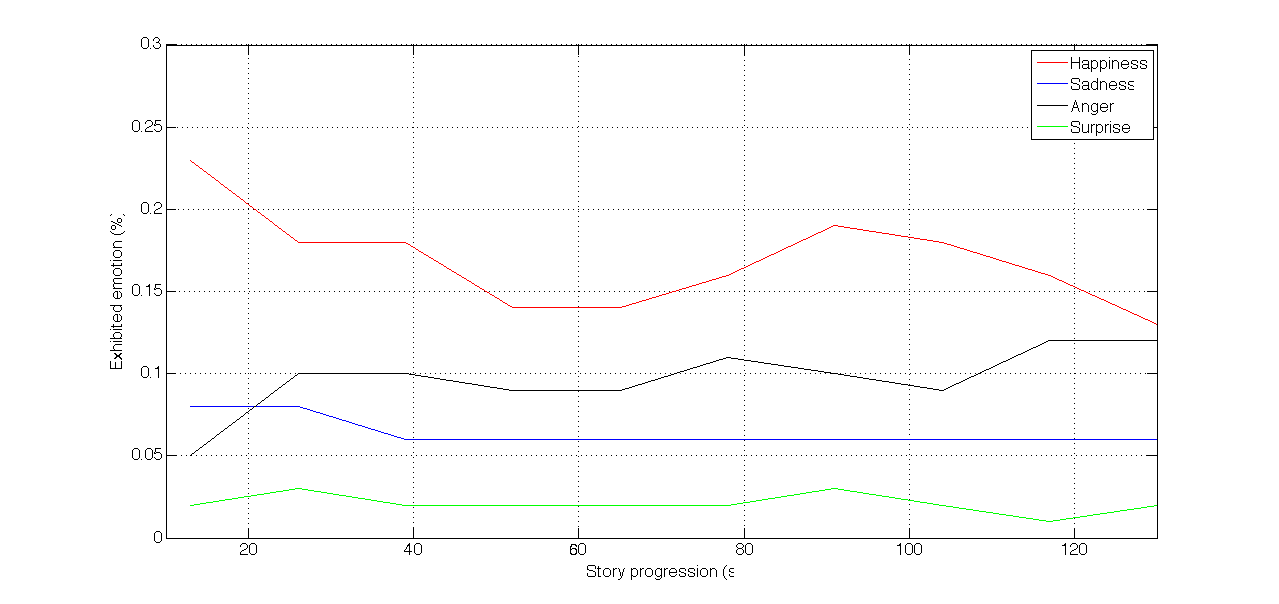}
	}
}
\caption{Percentage of the emotions shown in participants during storytelling by Aesop (Comparison between human and TTS voice).}
\label{Fig13}
\end{figure}

The comparison between the emotions displayed during storytelling with Aesop using a human voice and a TTS voice show \textit{significant differences} regarding anger (p \textless .001), sadness (p \textless .05), and surprise (p \textless .001). In average the percentage of the exhibited emotion regarding sadness and surprise was higher in the story told by the robot using a human voice than a TTS voice (Fig. \ref{Fig13}), thus illustrating the higher effectiveness of the human voice towards inducing emotions to the listener. \textit{This addresses our research question RQ2: indeed, in this respect a human voice seems to be more effective than the current state-of-the art text-to-speech computer voice towards storytelling agents}.

\subsubsection{Blinking}
\label{Blinking2}
Average blink rate decreased along the story, as was the case in Study 1. We then compared the blink frequency while listening to the story told by different agents. 
The story was again divided into three parts, and a paired sample t-test was used for comparison. 
We found \textit{significant differences} in the parts regarding the average of blinking rate (p \textless .05), 
when comparing the story being told by Greta and by Aesop. The average blinking rate is lower with Aesop than with Greta, 
indicating stronger listener engagement with the physical Robot as compared to the virtual Avatar, thus addressing our research question RQ1 - \textit{indeed, in this respect a physical robot seems to be more effective than an avatar as an emotional storyteller}.

\subsubsection{Body Language}
\label{BodyLanguage2}

\begin{figure}[tb]
\centering
\includegraphics[width=0.4\textwidth]{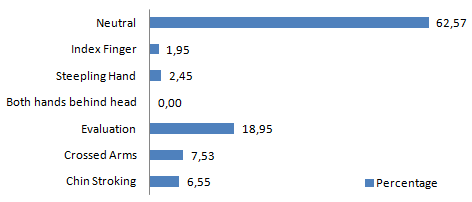}
\caption{Hand and arms gestures' percentages when the story was told by a) Greta and by b) Aesop.}
\label{Fig8_1}
\end{figure}

\begin{figure}[tb]
\centering
\includegraphics[width=0.4\textwidth]{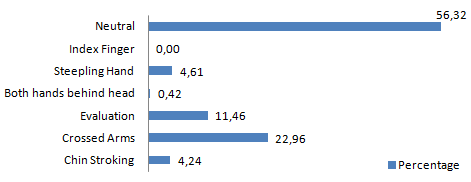}
\caption{Hand and arms gestures' percentages when the story was told by Aesop with a) a TTS voice and with b) a human voice.}
\label{Fig8_2}
\end{figure}

The videos from the experiments were used to evaluate the non-facial body language of the listener. Figs. \ref{Fig8_1} and \ref{Fig8_2} represents the gestures performed using hands and arms. In this category, the predominant gestures were: crossed arms, evaluation, chin stroking, steepling hands, and neutral. Crossed arms indicate that the person disagrees or is not comfortable with the situation in front of him/her, and evaluation is self implied. Chin stroking indicates the listener is making a decision, while steepling hands is an illustration of confidence \cite{pease2012body}. The ideal situation is that the participants' body language do not show behaviours indicating a negative attitude, as crossed arms, evaluation, chin stroking, and steepling hands.
\\ \indent ANOVA revealed \textit{significant differences} between the time percentage the participants performed the aforementioned behaviours when the story was told either by the Greta avatar or by the Aesop robot, F(6, 11) = 50.92; p \textless .0001. In fact, we verify that the participants have shown more neutral behaviours when listening to the story told by Aesop than when told by Greta. \textit{Significant differences} were also found comparing between the behaviours, what was expected already observing the percentage of time dedicated to neutral behaviours. \textit{Thus, in this respect too, regarding question RQ1, a physical robot seems to be preferable to a virtual agent as an emotional storyteller.}

When comparing the percentage of time the participants made these gestures when listening the story told by the robot either with a human voice 
or using a TTS voice, there are \textit{significant differences} F(6, 11) = 9.18; p \textless .01, and observing Fig. \ref{Fig8_2} 
we see that the story told using a human voice received more neutral behaviours. \textit{Regarding question RQ2, 
for a storytelling agent human voice seems to be preferable to TTS.}

Gestures associated with legs were then examined. Crossed legs normally occurs with other negative gestures, and leg clamp is a sign of the tough-minded individual. Having the legs parallel to each other is an indication of a neutral opinion \cite{pease2012body}. \textit{Significant differences} were found when comparing the percentage of time the participants performed legs gestures when the story was told either by Greta or by Aesop, F(2, 3) = 47296.49; p \textless .001 but no significant differences were found between the storytellers (p \textgreater .005). There were \textit{significant differences} between the percentage of time the participants made these gestures when listening the story told by the robot either with a human voice or using a TTS voice, F(2, 3) = 432.92; p = .002. Thus, what we found is that there is no significant differences between the storytellers in this behaviour. Regarding the leg movements of the participants, it is worth noting that given the positioning of the cameras and the table, in contrast to the movements of the rest of the body, the leg movements were neither always visible not were the legs absolutely free to move. However, we think it is worth reporting leg movements too, as the fact that some subjects chose to cross their legs even under these restrictive conditions might be a strong indication.

Regarding the head position category, we identified three different positions: neutral, interested and disapproval \cite{pease2012body}.
 A two-way ANOVA showed \textit{significant differences} between the percentage of time the participants showed the previously mentioned behaviours when the story was told either by Greta or by the robot, F(2, 3) = 49.48; p = .003, but no significant difference was found between the storytellers in this behaviour. The same is verified when comparing the percentage of time the participants made these gestures when listening the story told by the robot either with a human voice or using a TTS voice, F(2, 3) = 78,71; p = .01. Thus, we found no significant differences between the storytellers in this behaviour.
\\ \indent Thus, from the participants' non-verbal response analysis, we would like to highlight that the fact that the participants were seated in front of the storyteller having a table between them, did not allow the participants to freely exhibit behaviours from their legs. In addition, focus should be given to hands and arms gestures, where significant differences were found both between groups and between behaviours. 

To recap: significant differences across groups (robot vs. avatar and human voice vs. TTS) were found in the gestures involving head and arms, namely: crossed arms, evaluation, chin stroking, steepling hands, and neutral. These provided yet further results towards answering our questions RQ1 and RQ2, and supported the higher effectiveness of robot embodiment as compared to virtual avatars for storytelling, and the higher effectiveness of the human voice as compared to a synthesized voice for such storytelling.

\section{Discussion and Future Work}
\label{Discussion} 

Towards our ultimate goal of emotional storytelling agents that can observe their listener and adapt their style in order to maximise their effectiveness, we have set out three main research questions, which we investigated through experiments utilising questionnaires, automated analysis of facial expressions and blink rate, and manual analysis of non-verbal gestures.

The basic concern of Research Question 1 was \textit{embodiment}: is a physical robot preferable to a virtual avatar for emotional storytelling? Our results indeed indicate that it is. In particular, given our eye blink-rate as well as non-verbal gestures involving head and arms (crossed arms, evaluation, chin stroking, steepling hands, and neutral), we reach statistical significance twice providing strong support that physical robots are preferable for emotional storytelling.

As a sideways question, RQ1a, we asked whether there are differences in engagement across different parts of the story. As reported, blink rate data as well as questionnaire answers indicated that the apparent engagement was higher towards the climatic end of the story, attaining statistical significance.

Research Question 2 was concerned with the choice of storytelling \textit{voice}: is the state-of-the art of text-to-speech synthesisers good enough, or is a human voice still preferable? Our results indicate that a human voice is still preferable: both in terms of exhibited listener facial expressions indicating emotions during storytelling, as well as in terms of non-verbal gestures involving head and arms, we reached statistically significant difference twice. Still, text to speech technology has not reached the right level of maturity to compete with human voices for storytelling agents.

Finally, we reach the most interesting Research Question 3. Will empathy be exhibited by the listener's emotional reactions to the story? And at an even wider scope, a very important yet interesting question arises: what is the relation between the postulated emotional trajectories of different story characters, the emotional trajectory of the storyteller, and that of the listener? Is it a simple equality relation or a fixed one-to-one mapping? 

Let us start with empathy. Our results of Fig.4 and Fig.5 indicate that indeed there is empathy between the storyteller and the listener, as evidenced by the equality of valence between the narrator's facial expressions (corresponding to the dominant emotions of the story derived as described) and the listener's facial expressions, over time. Even more so, by looking at Fig.2 and Fig.3 and the commentary that follows them, it becomes evident that an interesting yet complex relation of emotional "harmonies": In analogy to musical harmony, where chords are formed through the vertical co-temporality of notes played by various voices, while the melodic lines of these voices are unfolding in parallel over time, one can postulate emotional "harmonies". I.e., as the emotional lines of different characters (story characters, storyteller, listener) evolve over time, one can postulate the vertical "chords" that are formed by them. As a simple example, one could imagine a situation where the listener has sympathised with a hero character, and the storyteller is acting out a villain character who is making an ironical happy statement to the hero. Then, the four emotional "voices" would be: (hero, villain, storyteller, listener). The resulting chord will be:

\begin{itemize}
\item 1st voice [Hero]: Sad (because of intimidation by villain)
\item 2nd voice [Villain]: Happy (making ironical statement to hero to intimidate)
\item 3rd voice [Storyteller]: Happy (acting out the villain's part)
\item 4th voice [listener]: Angry (because of sympathy to hero who is being abused)
\end{itemize}

Thus, the chord is:  Sad-Happy-Happy-Angry

As the story unfolds, many such chords follow one another, forming chord progressions. One could envision that there exist many interesting patterns, constraints, and relations, both on the temporal dynamics of the emotional lines of each voice (character, storyteller, listener), as well as on the vertical relations of the co-temporal content across the lines, and this opens up exciting avenues for further research.

For example, some such indications that we have observed in Fig. 2 and Fig. 3 indicate that there exist cases where there is equality of voices, such as event E3(happy) = E3'(happy), and other cases where there is no simple fixed 1-1 relation E1(sadness) - E1'(anger), while E2(sadness) - E2'(anger and sadness). Of course, one can postulate explanation behind these patterns, in terms of both the temporal progression of the story (horizontal aspect) as well as the co-temporal reactions to emotional states. For example, it might be the case that during E1, while the narrator is sad because of the sadness of the hero (cracked pot), the listener feels angry because of the perceived injustice and sympathy for the hero who has to be sad without the reason being his personal fault. Later, in E2, as story time goes on and the situation is not reversed, the listener's anger falls and sadness increases. But when finally at E3, the good news is announced, that the crack was not only not the pot's fault but that it has produced so many unique flowers and that the master is happy, then quickly both the storyteller as well as the listener become happy: and the listener leaves with an almost persistent smile on their face. This is a plausible account of the observed emotional melodies and harmonies of the storyteller and the listener, and of course there as just the beginnings of a potential theory connecting concepts of music (melody, harmony) with emotional trajectories of multiple characters evolving over time. 

Another interesting comment is the following. The results from the questionnaires in Study 2 (section \ref{Questionnaires2}) 
suggest the participants were empathising less with the main character in the beginning of the story (Q2), 
when it was feeling bad about itself. However, Q5 scores show us that in the end of the story the participants 
shared its joy more strongly, and as we discussed above, one possible explanation is that an empathetic connection 
between the listener and the hero was being formed and strengthened along the story. This is a good indication that 
the way the story was encoded in the Study 1 promoted empathy. In the first part of the story, the listener might had been 
building a mental model of the story and its entities, and establishing a connection with the storyteller and the characters. 
While the situation model of the story was being created in the listener's mind \cite{zwaan1998situation} \cite{mavridis2007}, 
the mental models of the characters represented in the listener's mind started as generic models of human characters, without specific 
information attached to them. However, the more the listener learned about them as the story narration progressed, they started becoming 
more specific. For stronger empathising, the story needed to have progressed enough in order to know enough about the character in order to 
empathise strongly with him or her, and with his traits, and history. This result is congruent with the findings of RQ3, 
which indicate empathy at the valence level between storyteller and listener, again supporting the effectiveness of our storytelling agent.

Numerous future extensions exist: First, in the current study, a single story was only used. 
A small yet varied repertoire of stories of similar duration would be beneficial, 
and the variation could help obtain more cases of interesting emotional line progressions and harmonies. 
Furthermore, if one extends to longer stories, than other phenomena, such as total loss of attention, might become apparent. 
Also, we could start modulating not only the facial expressions of the storytelling agent, but also his voice prosody, 
temporal parameters such as pauses etc. 
Furthermore, the level of the anthropomorphicity of the storyteller could be varied, and also the age groups of listeners. 
Most importantly, the above notes on the emotional "harmonies" and "melodies" that take place during storytelling across characters, 
storyteller, and listener, could be turned into a formal theory, and targeted experiments performed to further inform it.


\section{Conclusion}
\label{Conclusions}

Towards our ultimate goal of real-time adaptive emotional storytelling agents, we have presented experiments and results towards answering three important research questions, termed RQ1-RQ3, and we have furthermore provided a discussion including the early steps of a theory of emotional "melodies" and "harmonies" in analogy to their musical counterparts.

We started by noting that numerous aspects of narration, including facial expressions and prosody, are important towards creating interesting and effective artificial storytellers, either robotic or virtual. Then, we noted that ideally, the manner of narration should be adaptable in real-time during storytelling, based on feedback derived from non-verbal cues arising from the listener, given the differences in engagement and preferences of different listeners at different times. But in order to start experimenting with either off-line or on-line real-time adaptation, a number of fundamental research questions came up, whose answers are highly important: First, is a physically embodied agent preferable to a virtual agent or a voice-only narration? Second, does a human voice have an advantage over a synthesized voice? Third, how should the emotional trajectory of the characters in a story be related to a storyteller's facial expressions during storytelling time, and how does this correlate with the emotions on the faces of the listeners? 

In this paper, we provided empirical answers to the above questions, through two specially designed studies, during which we observed the reactions of experimental subjects to artificial storytellers, through a combination of instruments: special questionnaires, manual body language annotation, and automated facial expression and blink analysis. The results indicate that the physically embodied robot produces more attention to the listener as compared to a virtual embodiment, that a human voice is preferable over the current state of the art of text-to-speech, and that the empathising of the listener is evident through its facial expressions. Most importantly, it became apparent that there is a complex yet interesting relation between the emotion lines of the story, the facial expressions of the narrating agent, and the emotions of the listener, and the beginnings of a theory of emotional "melodies" across time and "harmonies" agents were introduced in our discussion, where further future steps were also discussed. This work constitutes an important step towards emotional storytelling robots that can observe their listener and adapt their style in order to maximise their effectiveness, thus enabling the further beneficial entry of robots towards enhancing our everyday life.

\section*{Acknowledgements}
To the IRSS2013 participants, to the Portuguese Foundation (FCT) for funding through the R\&D project RIPD/ADA/109407/ 2009, 
SFRH/BD/71600/2010, and FCOMP-01-0124-FEDER-022674, to the Brain Korea 21 Plus Program through the NRF funded by the 
Ministry of Education of Korea (10Z20130000013).

\newcommand{\BIBdecl}{\setlength{\itemsep}{0 em}}

\bibliographystyle{IEEEtran}
\bibliography{IRSS13bib}

\begin{IEEEbiographynophoto}{Sandra Costa} received her Msc degree by the Minho University in Electronic Engineering in 2009, focusing her research in using a mobile and modular robot to interact with individuals with autism. At the moment, she is a Ph.D. student in the same university investigating how can a humanoid robot develop socio-emotional skills in children with autism. Since 2010 she is working in Algoritmi Center and her main scientific interests are affective robotics, human-robot interaction, and robotics used for rehabilitation.
\end{IEEEbiographynophoto}

\vspace*{-2\baselineskip}

\begin{IEEEbiographynophoto}{Alberto Brunete}
received a M.Sc. degree in Telecommunications and a Ph.D. degree in Robotics and Automation from the Universidad Polit\'ecnica de Madrid (UPM), where he works as an Assistant Professor and researcher (CAR UPM-CSIC). He has been visiting professor at the Carlos III University. He has work as technical coordinator for the Research Centre for Smart Buildings and Energy Efficiency (CeDInt-UPM). His main research activities are related to robotics, smart environments and Internet of things.
\end{IEEEbiographynophoto}

\vspace*{-2\baselineskip}

\begin{IEEEbiographynophoto}{Byung-Chull Bae}
Byung-Chull Bae received the BS (1993) and the MS (1998) degree in Electronics Engineering from Korea University, South Korea, and the Ph.D. degree in Computer Science from North Carolina State University, Raleigh, NC in 2009. He has worked at LG Electronics and Samsung Electronics as a research engineer, and worked for IT University of Copenhagen in Denmark as a visiting scholar and an external lecturer. He is currently Assistant Professor at School of Games, Hongik University in South Korea. His research interests include interactive narrative, affective computing, and games.
\end{IEEEbiographynophoto}

\vspace*{-2\baselineskip}

\begin{IEEEbiographynophoto}{Nikolaos Mavridis}
Nikolaos Mavridis received his Ph.D. from MIT in 2007, after receiving his M.S.E.E. from UCLA and a M.Eng. in ECE from the Aristotle University of Thessaloniki. 
He is the founder and director of the Interactive Robots and Media Lab (IRML), and has served as Assistant Professor at UAEU, NYU AD, and currently is with NCSR Demokritos. 
The research interests of Dr. Mavridis include Human-Robot Interaction, Computational Intelligence, and Cognitive Systems.
\end{IEEEbiographynophoto}

\end{document}